\documentclass[11pt,a4paper]{article}
\usepackage[hyperref]{acl2020}
\usepackage{times}
\usepackage{latexsym}
\usepackage{graphicx} 
\usepackage{amsmath}
\usepackage{url}
\usepackage{hyperref}

\usepackage{parskip}
\usepackage{multirow}
\usepackage{graphics}
\usepackage{parskip}
\usepackage{amssymb}
\usepackage{amsmath}
\usepackage{float}
\usepackage{enumitem}
\usepackage{booktabs}
\usepackage{xspace}
\usepackage{makecell}
\usepackage{microtype}

\aclfinalcopy 

\title{An Overview of Distant Supervision for Relation Extraction with a Focus on Denoising and Pre-training Methods}

\author{
William P Hogan \\
Department of Computer Science \& Engineering\\
University of California, San Diego \\
}

\date{}

\begin{document}
\maketitle

\begin{abstract}
Relation Extraction (RE) is a foundational task of natural language processing. RE seeks to transform raw, unstructured text into structured knowledge by identifying relational information between entity pairs found in text. RE has numerous uses, such as knowledge graph completion, text summarization, question-answering, and search querying. The history of RE methods can be roughly organized into four phases: pattern-based RE, statistical-based RE, neural-based RE, and large language model-based RE. This survey begins with an overview of a few exemplary works in the earlier phases of RE, highlighting limitations and shortcomings to contextualize progress. Next, we review popular benchmarks and critically examine metrics used to assess RE performance. We then discuss distant supervision, a paradigm that has shaped the development of modern RE methods. Lastly, we review recent RE works focusing on denoising and pre-training methods.
\end{abstract}

\section{Introduction} \label{sec:introduction}
Relation extraction (RE), a subtask of information extraction, is a foundational task in natural language processing (NLP). The RE task is to determine a relationship between two distinct entities from text, producing fact triples in the form [\textit{head}, \textit{relation}, \textit{tail}] or, as referred to in some works, [\textit{subject}, \textit{predicate}, \textit{object}]. For example, after reading the Wikipedia page on Noam Chomsky, we learn that Noam was born in Philadelphia, Pennsylvania, which corresponds to the fact triple [\textit{Noam Chomsky}, \textit{born in}, \textit{Philadelphia}]. Fact triples are foundational to human knowledge and play a key role in many downstream NLP tasks such as question-answering, search queries, and knowledge-graph completion \cite{Xu2016QuestionAO, Lin2015LearningEA, Li2014CoREAC}.

\begin{figure}[t]
    \centering
    \includegraphics[width=0.47\textwidth]{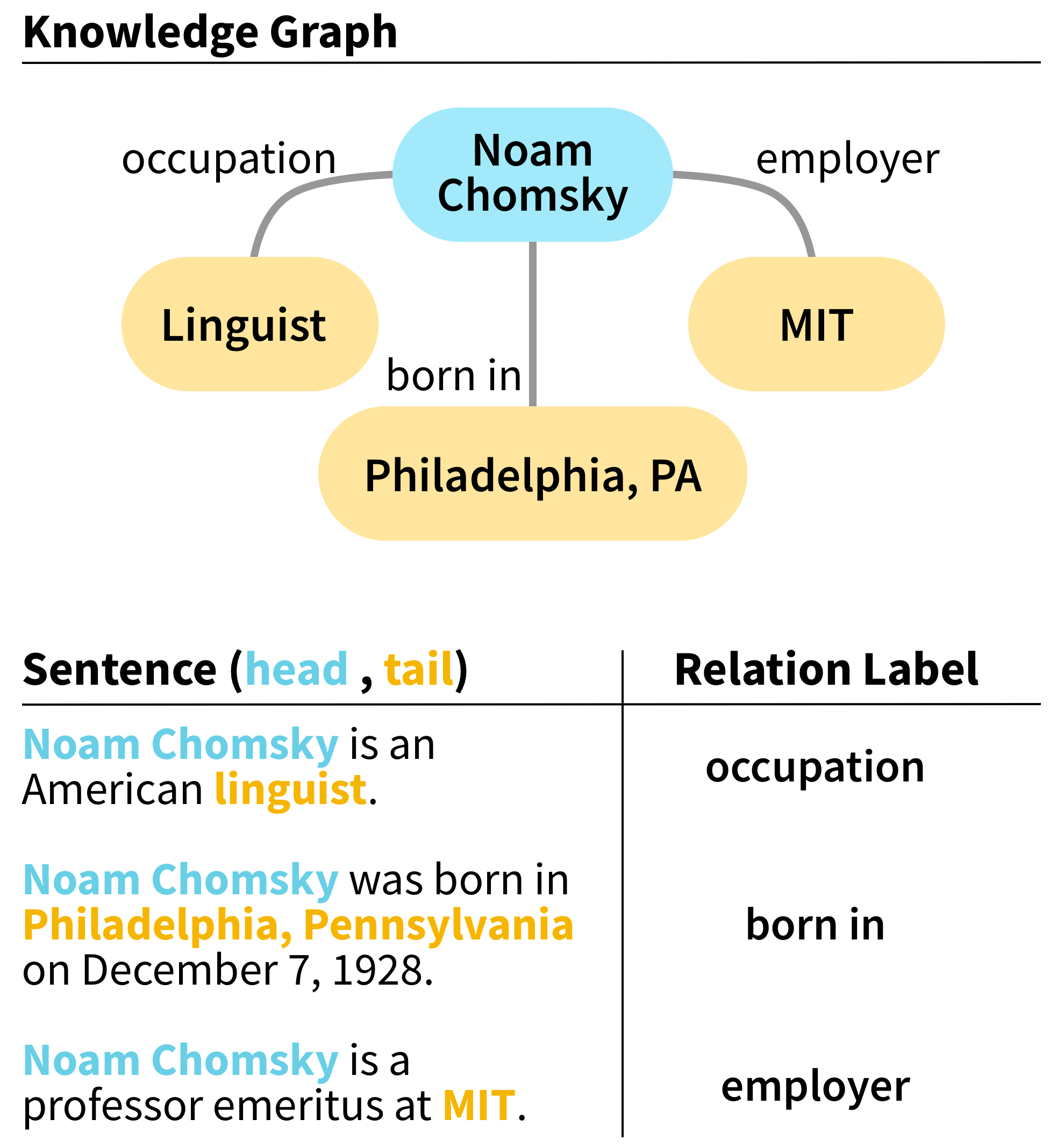}
    \caption{A sample of relation labels generated by distant supervision. A knowledge graph, here a small snippet from WikiData \cite{wikidata}, is paired with sentences to label instances of relations linking an entity pair.}
    \label{fig:re_sample}
\end{figure}

Distant supervision for relation extraction is a method that pairs a knowledge graph---a graph of entities connected by edges labeled with relation classes---with an unstructured corpus to generate labeled data automatically \cite{Mintz2009DistantSF}. First, entities from the knowledge graph are identified in the text, and then the following assumption is made: all sentences containing an entity pair express the corresponding relation class, as determined by the accompanying knowledge graph. Figure \ref{fig:re_sample} provides an example of automatically generated relation labels pairing a knowledge graph, namely WikiData \cite{wikidata}, with raw sentences.

Formally, the problem statement for distantly supervised relation extraction is as follows: given knowledge graph $\mathcal{G}$ and text corpus $\mathcal{C}$, sentences $s$ from $\mathcal{C}$ that contain two distinct entities, $({e^i}_1, {e^i}_2)$, that are linked via relationship $r^i$ as determined by $\mathcal{G}$, where ${e^i}_{1}, {e^i}_{2}, r^i \in \mathcal{G}$,  predict the relationship $r^i$ that is expressed in sentence $s^i$, forming a fact triple $t = \{({e^i}_1, {r^i}, {e^i}_2)\}$.

This survey is organized as follows:
\begin{itemize}
    \item In Section \ref{sec:background}, we conduct a high-level overview of relation extraction. We trace the history of RE methods and highlight limitations and shortcomings to contextual progress. We review both how and why distant supervision for RE was developed. 
    \item In Section \ref{sec:datasets}, we review popular RE datasets and critically examine evaluation metrics used to assess RE performance. 
    \item In Section \ref{sec:main}, we review recent methods used for distantly supervised RE. We focus on the two primary methods characteristic of distantly supervised RE: denoising methods and pre-training methods.
\end{itemize}
\section{Background}\label{sec:background}
The history of RE methods can be roughly organized into four phases: (1) pattern-based RE, (2) statistical-based RE, (3) neural-based RE, and (4) large language model-based RE. We provide an approximate year when a particular phase begins, but we do not provide an end year as many of the methods from earlier RE phases continue into subsequent phases, creating a rich blend of techniques and no clear end to any particular phase. 

\subsection{Pattern-based RE ($\sim$1970\texttt{+})} 
As the name suggests, pattern-based RE is the development of algorithms to learn linguistic patterns to extract relation information from text. Pattern-based RE methods typically use seed words or phrases combined with knowledge sources such as a part–of–speech tagger and semantic class information to identify patterns within text relevant to relation extraction.

In \citet{Califf1997RelationalLO}, an exemplary work of pattern-based RE, the authors propose the ``Robust Automated Production of Information Extraction Rules'' (RAPIER) algorithm. RAPIER is a form-filling algorithm that uses form fields as seed words to extract relational information. The algorithm expands the seed words to include all synonyms via the synsets provided by WordNet \cite{wordnet}, a large lexical database with extensive synsets for English words. Words within sentences are tagged with a part-of-speech (POS) tagger and RAPIER then attempts to learn patterns of POS tags before and after seed words within a sentence. The authors develop and evaluate RAPIER on a small dataset of job listings, extracting facts such as job location, required skills, and salary.

Pattern-based methods typically involve bespoke algorithms developed and refined within a specific domain, and RAPIER is no different. Because of this, pattern-based methods often fail to generalize to out-of-domain data. Moreover, while pattern-based methods typically achieve high precision on relation extraction benchmarks, they suffer from low recall. These limitations motivated the development of statistical-based relation extraction methods. 

\subsection{Statistical-based RE ($\sim$2004\texttt{+})} 
In the early 2000s, advanced statistical-based RE methods gained popularity. Statistical-based RE looked to leverage advances in machine learning to extract relations from text. Hand-crafted textual features, such as dependency trees and part-of-speech tags, were typically used to train logistic regression classifiers.

Statistical-based machine learning methods require large datasets for training and evaluation. However, manually labeling relations in text is both expensive and time-consuming. Dataset creators must develop annotation guidelines to ensure consistency and pay and train human annotators (e.g., via Mechanical Turk \cite{Buhrmester2011AmazonsMT}) to annotate data. 

To address the data scarcity issue, \citet{Mintz2009DistantSF} released a pivotal work titled ``Distant supervision for relation extraction without labeled data.'' In it, the authors coin the term \textit{distant supervision for relation extraction} and define it as the assumption that any sentence containing an entity pair may express a relationship between that entity pair as determined by an accompanying knowledge graph. This simple but powerful assumption unlocks the ability to automatically generate the large amounts of training data needed for advanced statistical RE models. 

In the original distant supervision for RE work, \citeauthor{Mintz2009DistantSF} pair the Freebase \cite{Bollacker2008FreebaseAC} knowledge graph with text from Wikipedia (English)\footnote{\url{https://en.wikipedia.org/}} and, in doing so, generated a dataset consisting of 1.8 million instances of 102 relation classes connecting 940,000 entities. They trained a multi-class logistic regression classifier with numerous hand-crafted lexical and semantic features using this data. For lexical features, they use the sequence of words between two entities, POS tags, a binary flag indicating which entity in a pair is mentioned first in the sentence, a window of $k$ words and POS tags to the left of the first entity as well as words and POS tags to the right of the second entity. For semantic features, they include word dependencies via the dependency parser MINIPAR \cite{Lin2003} and named entity tags via the Stanford four-class named entity tagger \cite{Finkel2009NestedNE}. With the help of human annotators, they observed that their logistic regression classifier, leveraging distant supervision and a combination of hand-crafted features, was able to extract fact triples with decent precision (0.67 on the highest-ranked 1000 results per relation). 

Notably, most of the early distantly supervised RE work focused on intra-sentence relations (i.e., ``sentence-level'' relations). However, \citet{Quirk2017DistantSF} observed that many relations within a document were expressed across multiple sentences (i.e., cross-sentence, inter-sentence, or ``document-level'' relations). To extract document-level relations, the authors proposed a statistical-based RE method named DIstant Supervision for Cross-sentence Relation EXtraction (DISCREX). DISCREX first labels words with their corresponding lemma and part-of-speech. It then forms a document graph of words by leveraging Stanford’s dependency parsing tool which outputs dependency paths between words in a sentence. The model searches within the document graph to find the shortest path between two entity mentions, ignoring sentence boundaries. The intuition is that a relationship between an entity pair is more likely to be expressed when entities are closer together, regardless of sentence boundaries. The authors train DISCREX to detect an association between an entity pair via binary logistic regression. DISCREX is notable because it is an early model focusing on document-level RE. As discussed in Section \ref{sec:datasets}, document-level RE has arguably become the primary focus for current RE methods \cite{BaldiniSoares2019MatchingTB, Peng2020LearningFC, Qin2021ERICAIE}.

\subsection{Neural-based RE ($\sim$2010\texttt{+})} \label{sec:neural-based-re}
In the early 2010s, with help from large automatically labeled datasets enabled by distant supervision for RE and the advance of neural networks, RE entered its third phase---neural-based RE. Neural-based RE departs from statistical-based RE in two key ways: (1) by using a neural network instead of logistic regression, and (2) by replacing hand-crafted textual features with textual features learned via a neural network (e.g., Word2Vec \cite{NIPS2013_9aa42b31} and GloVe \cite{pennington-etal-2014-glove}). 

Neural RE models come in a variety of architectures, such as the recurrent neural network (RNN) \cite{Zhang2015RelationCV}, long-short term memory (LSTM) network \cite{Peng2017CrossSentenceNR, Miwa2016EndtoEndRE}, and the convolutional neural network (CNN)\cite{Zeng2015DistantSF}. During this era, RE methods enjoyed a significant boost in performance.

Relationships in text may be expressed within a long sentence or across multiple sentences and, while high-performing, many early neural-based models struggled to learn such long-range dependencies. To address this shortcoming, \citet{Vaswani2017AttentionIA} introduced the transformer architecture, a seminal work that \textit{transformed} RE and NLP in general. 

The transformer is similar in structure to other sequence transduction models in that it consists of two modules: an encoder module and a decoder module. The encoder takes an input sequence $\textbf{x}$ and produces a dense representation $\textbf{z}$ that is fed to the decoder. The decoder then uses $\textbf{z}$ to produce an output sequence $\textbf{y}$. However, the transformer was unique in its use of stacked multi-head attention functions in the encoder and decoder modules. 

In the original transformer architecture, the encoder and decoder modules consisted of six identical and serially-connected layers. Each layer contained a multi-head attention function as well as a feed-forward network. The initial encoder layer is fed an input sequence $\textbf{x}$, and, using multi-head attention, it attends to all positions in that sequence---that is, it learns which parts of the input sequence to focus on for a given task. Next, the result from the multi-head attention function is sent through a feed-forward network and, finally, to the next layer in the module. Subsequent layers repeat this process using the previous layer's output as their input. The transformer is arguably the most impactful neural-based architecture for RE and NLP as it is the primary architecture used in large pre-trained language models.

\subsection{Large language model-based RE ($\sim$2019\texttt{+})} 
The transformer architecture inspired the development of BERT \cite{Devlin2019BERTPO}, a wildly effective pre-trained language model that produces word embeddings richer and more informative than Word2Vec and GloVe. BERT's success inspired a slew of other pre-trained models, also known as foundation models \cite{foundation_models}, all rooted in the same two-step training paradigm---a large, self-supervised pre-training followed by a smaller, supervised fine-tuning. This training methodology brought RE into its fourth and current phase. 

By leveraging pre-trained language models, RE performance improved across all benchmarks. At the time of writing, the top ten models on the DocRED dataset \cite{docred}, the most challenging general domain RE benchmark, all leverage BERT or a BERT relative (e.g., RoBERTa \cite{Liu2019RoBERTaAR})\footnote{\url{https://paperswithcode.com/sota/relation-extraction-on-docred}}. We conduct a deeper dive into these leading models in Section \ref{sec:main}.

\section{Datasets \& Benchmarks}\label{sec:datasets}
This section briefly introduces some popular RE datasets and benchmarks used in many of the works discussed throughout this survey. We also critically examine the current evaluation metrics used to assess RE performance.

The development of relation extraction datasets mirrors the history of RE development. First, small, human-labeled datasets were developed for pattern-based methods, followed by larger manually and automatically labeled datasets used to train statistical models and neural networks \cite{Riedel2010ModelingRA, zhang2017tacred, docred}.

\begin{table*}[t]
  \centering
  \begin{tabular}{ l | c | c | c | c } \hline
    Dataset             & RE Type           & Data Creation Method  & Instances & Classes \\ \hline \hline
    SemEval-2010 Task 8 & Sentence-level    & Manual                & 8.8k      & 9     \\ \hline
    NYT10               & Sentence-level    & Automatic             & 21.8k     & 53    \\ \hline
    TACRED              & Sentence-level    & Manual                & 106k      & 41    \\ \hline
    FewRel              & Sentence-level    & Manual                & 70k       & 100   \\ \hline
    BC5CDR              & Doc-level         & Manual                & 3.1k      & 1     \\ \hline
    DocRED              & Doc-level         & Auto + Manual         & 1.56M     & 96    \\ \hline
  \end{tabular}
  \caption{A sample of sentence-level and document-level RE datasets highlighting the dataset creation method, size, and the number of relation classes. Manually created datasets are created by human annotators, while automatically generated datasets are created using distant supervision for relation extraction.}
  \label{table:datasets}
\end{table*}

RE datasets typically focus on one of two sub-tasks: (1) sentence-level RE or (2) document-level RE: 

\textbf{1. Sentence-level RE} seeks to extract relationships between entity pairs within the bounds of a single sentence (i.e., intra-sentence relationships). Some popular sentence-level RE datasets include TACRED \cite{zhang2017tacred}, NTY10 \cite{Riedel2010ModelingRA}, SemEval-2010 Task 8 \cite{hendrickx-etal-2010-semeval}, and FewRel \cite{han-etal-2018-fewrel, gao-etal-2019-fewrel}. 

Sentence-level RE is criticized for being overly simplistic since not all relationships in text are expressed within a single sentence. \citet{docred} report that 40.7\% of relationships in Wikipedia documents are expressed across multiple sentences. That is, to identify 40.7\% of instances of relationships in Wikipedia, a human or machine learning model must comprehend multiple sentences to determine a relationship between an entity pair.

\textbf{2. Document-level RE} was developed to address the shortcoming of sentence-level RE by extracting relations between entity pairs contained within a document (i.e., both inter- and intra-sentence relationships). In the general domain, the DocRED dataset \cite{docred} is a popular document-level RE dataset that pairs text from Wikipedia to entities and relations from the WikiData knowledge graph \cite{wikidata}. In the biomedical domain, the BioCreative V CDR task corpus (BC5CDR) \cite{bcvcdr} is a popular document-level dataset containing binary associations between chemical and disease entities using text from PubMed \cite{canese2013pubmed}.

Table \ref{table:datasets} lists common RE datasets, noting the RE type (sentence-level versus document-level), the creation method (manually versus automatically labeled), dataset size, as well as the number of relation classes they contain, not including ``no relation.''

\subsection{Evaluation}\label{sec:evaluation}
There are two common ways to evaluate a model's ability to extract relationships: \textbf{corpus-based} and \textbf{instance-based}.

\textbf{Corpus-based evaluation}, also known as ``hold out'' evaluation \cite{Mintz2009DistantSF}, evaluates a model's ability to predict an unseen, or ``held out,'' set fact triples in a test corpus. It is more commonly used for distantly supervised data that lacks gold labels. For corpus-based evaluation, precision-recall (PR) curves, area under the PR curve (AUC), and precision@$k$ is reported. Precision@$k$ is generated by pooling all of the model's inferred predictions on a test set, sorting them by their softmax probabilities, and then reporting the precision for the top $k$ predictions. 

\textbf{Instance-based evaluation} assesses a model's ability to predict each relation instance in a test set. For this, standard precision, recall, and F1-Micro score are used. Instance-based evaluation is commonly used on manually annotated datasets where all instances of relations have gold labels.

\subsubsection{Evaluation Limitations}
Both corpus-based and instance-based evaluations have limitations. Corpus-based evaluation suffers from false negatives since the held-out set of fact triples, generated from the paired knowledge graph, may be incomplete. Furthermore, in \citet{Gao2021ManualEM}, the authors manually examined distantly supervised relation data and found that as many as 53\% of the assigned labels were incorrect. They argue that, with such a high percentage of incorrectly labeled data, automatic corpus-based evaluation is not a representative metric of model performance. They suggest that distantly supervised datasets must be paired with and evaluated on manual labels. 

One shortcoming of instance-based evaluation is that it rewards the memorization of triples seen in training. Some benchmarks, such as those used for DocRED, avoid this biasing by reporting an F1 score calculated only on unseen triples in addition to an overall F1 score. 

Another shortcoming of instance-based evaluation is that a model's performance is boosted by correctly predicting multiple instances of common relation classes. This can obscure performance on rare, long-tail instances. In \citet{Gao2021ManualEM}, the authors note that many RE datasets have long-tail distributions of relation classes, making it possible for an RE model to achieve high F1-Micro scores by performing well on common classes and poorly on rare, long-tail classes. They compare a few leading RE models on the NTY10 data \cite{Riedel2010ModelingRA} and show almost no difference in performance on the top four relation classes in the dataset. There are, however, significant differences in performance on less-common relation classes. This is because less-common classes have fewer supporting sentences, making contextual learning difficult. In these cases, models that make predictions based on shallow knowledge graph heuristics perform well. Yet, such models can produce large amounts of false positives---predicting relationships in sentences without supporting context. As such, \citeauthor{Gao2021ManualEM} suggest adopting F1-Macro scores, in addition to the F1-Micro score, for instance-based evaluation to highlight model performance on long-tail relation classes. 

Using corpus-based and instance-based evaluation, neural models have demonstrated superior performance, yet what they are actually learning remains under-explored. For instance, it is unclear whether a model leans more on entity mentions or supporting context to make its predictions. Ideally, a model extracts relationships by understanding both the entity mentions and the supporting context, but a recent work shows otherwise \cite{Peng2020LearningFC}.

In \citet{Peng2020LearningFC}, the authors use the TACRED dataset and create five experimental settings designed to determine whether context or entity mentions are more informative in leading RE models. In the first setting, ``context and entity mentions,'' they train models using unaltered entity mentions and context. This is the baseline setting used in almost all RE models. In the second setting, ``context and entity types,'' they remove entity mentions from sentences and replace them with the corresponding entity types. For instance, the entity mention ``Angela Davis…'' is replaced with its entity type, namely ``[person].'' This setting is designed to determine the impact of entity type information in model performance relative to the default surface forms of entity mentions. The third setting, ``only context,'' masks entity mentions entirely and has the model make predictions based purely on context. The fourth setting, ``only entity mentions,'' masks the context and only allows the model to make predictions based on entity mentions. Finally, in the fifth setting, ``only entity types,'' they mask context and replace entity mentions with the corresponding entity types. 

The best performance resulted from the ``context and entity types'' setting where entities in a sentence were replaced by the corresponding entity types, indicating that type information from entity mentions is more informative to RE models than the entity mentions themselves. ``Only context'' outperformed ``only mentions'' which shows that models learn more from context than entity mentions. The authors conducted an error analysis on the ``only context'' results and found that 43\% of incorrectly predicted instances contained sufficient context to make a correct prediction, highlighting that current models can improve by better learning context.

Interestingly, while the ``only context'' setting outperformed ``only mentions'' setting, the ``only mentions'' setting still achieved good overall RE performance. That is, with no context at all, models could still predict relations between entities. This highlights how pre-trained language models contain some latent understanding of entities stored in their parameters. This can be problematic---pre-trained models may be prone to making predictions based on shallow heuristics, predicting a relationship between entities despite a lack of supporting textual context. This leads to many false positives and inflated performance on few-shot relation instances \cite{Peng2020LearningFC}, which matches the observation made by \citet{Gao2021ManualEM}.

Considering these findings, the authors suggest that future RE works focus on training models that prioritize learning relationships from context and reduce biasing from shallow heuristics. Additionally, RE evaluation metrics may benefit from reporting performance on instances with and without sufficient context. This will help reveal which models effectively learn context versus models that predict relations based on shallow knowledge graph heuristics.

\section{Current Methods}\label{sec:main}
This section delves into recent influential works on distant supervision for relation extraction. We divide the discussion into two primary methods characteristic of current distant supervision for RE works: \textbf{denoising methods} and \textbf{pre-training methods}.

\subsection{Denoising Methods}
\begin{figure}
    \centering
    \includegraphics[width=0.48\textwidth]{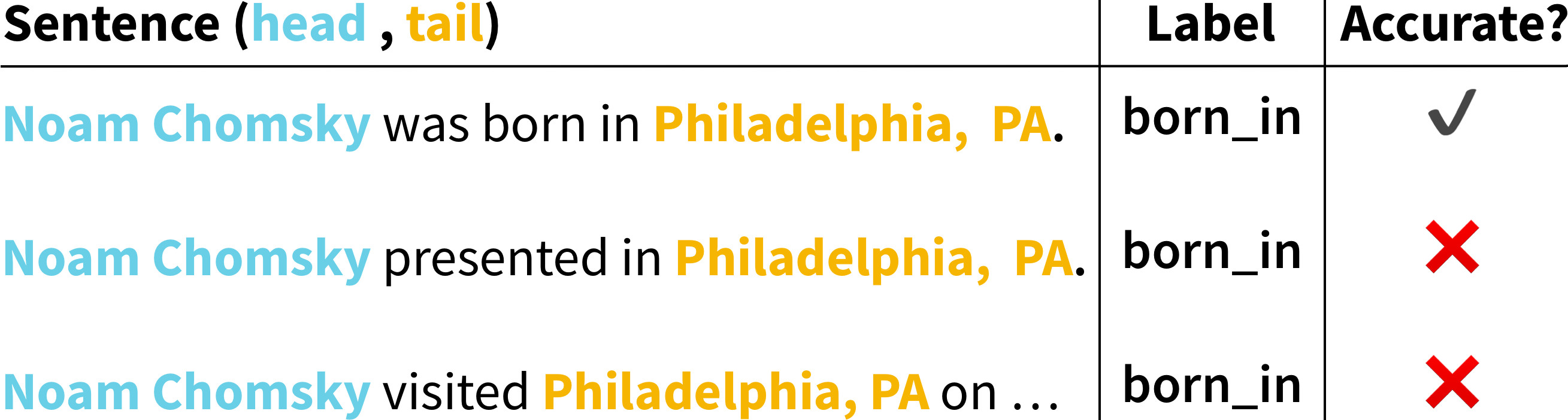}
    \caption{A sample of the wrong-labeling issue by automatically labeled sentences using distant supervision for relation extraction. Only the first sentence accurately expresses the labeled relationship ``born in.''}
    \label{fig:wrong_labels}
\end{figure}

As discussed in Section \ref{sec:introduction}, distant supervision for relation extraction pairs a knowledge graph with raw textual data and assumes that any sentence that contains two entities linked in the knowledge graph expresses the corresponding relationship. However, using this method to automatically label data produces a noisy training signal in the form of false positives since not all sentences will express a relationship. For example, consider the sentences in Figure \ref{fig:wrong_labels}. Using distant supervision for RE, each sentence is labeled as a positive instance of the relationship ``born in,'' however, only the first sentence properly expresses the relationship. As such, a large portion of the work that leverages distant supervision focuses on developing advanced denoising methods. 

\begin{figure}
    \centering
    \includegraphics[width=0.48\textwidth]{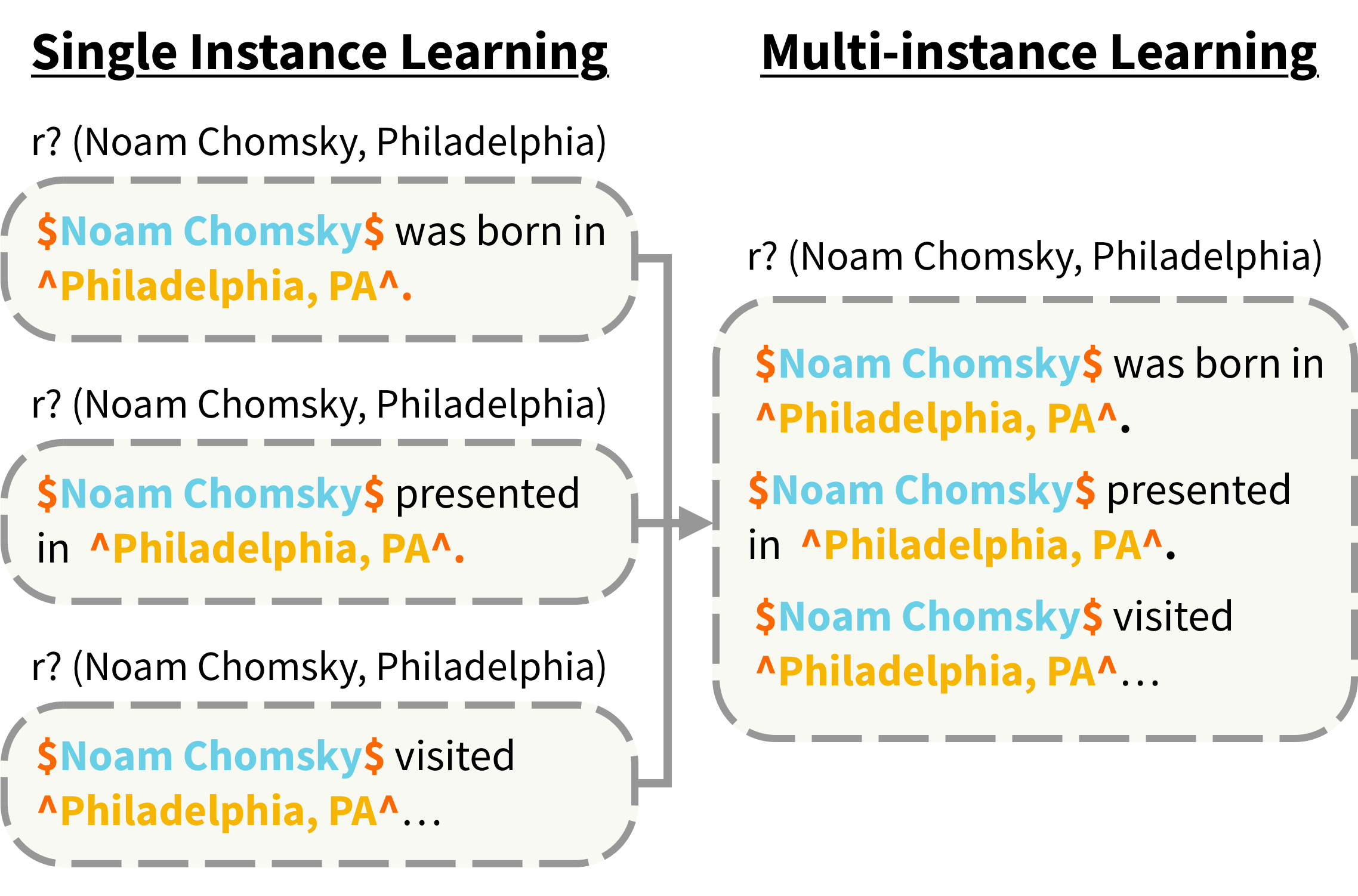}
    \caption{A sample of single-instance learning (left) and multi-instance learning (right) proposed by \citet{Riedel2010ModelingRA}. Multi-instance learning is an effective denoising method for distantly supervised RE as a group of sentences will more likely express a relationship than a single sentence. In this example, head and tail entities are wrapped in special tokens (`\textbf{\$}' and `\textbf{\^}', respectively), following a method popularized by \citet{BaldiniSoares2019MatchingTB}.}
    \label{fig:mil}
\end{figure}

Shortly after distantly supervised RE was proposed by \citet{Mintz2009DistantSF}, \citet{Riedel2010ModelingRA} proposed ``multi-instance learning'' (MIL) as a more relaxed version of the original distant supervision assumption designed to better handle noisy labels. Instead of assuming every sentence containing an entity pair expresses a relationship, multi-instance learning, as shown in Figure \ref{fig:mil}, collects a group, or ``bag,'' of sentences that contain the same entity pair and assumes that \textit{at least one} sentence in the group of sentences expresses the corresponding relationship. The intuition supporting multi-instance learning is simple: when considering sentences in a corpus that share the same entity pair, a group of those sentences are more likely to express a relationship than any single sentence.

In \citet{Surdeanu2012MultiinstanceML}, the authors expand on multi-instance learning. They note that some entity pairs have multiple relationships; however, there is no way to determine the best relationship label to apply when generating labels using distant supervision. Analyzing the NTY10 dataset \cite{Riedel2010ModelingRA} they report that multiple relationships connect 7.5\% of entity pairs.

To address this issue, \citeauthor{Surdeanu2012MultiinstanceML} propose a multi-instance, multi-label model for RE (\textsc{miml-re}). \textsc{miml-re} models multiple relationships via an Expectation Maximum (EM) algorithm that learns a single latent representation, $\textbf{z}$, which expresses $k$ relation classes.  They feed $\textbf{z}$ into a multi-label classifier that is trained to predict the relationship expressed between an entity pair. The multi-label classifier can learn which relational classes can and cannot jointly occur. For example, relations ``\textit{capital of}'' and ``\textit{contained}'' can jointly occur within a single entity pair while ``\textit{born in}'' and ``\textit{spouse of}'' cannot. Tested on NYT10 and compared to the original multi-instance model proposed by \citet{Riedel2010ModelingRA}, they report their \textsc{miml-re} model increases precision by 2 to 15 points at the same recall point. 

\citeauthor{Riedel2010ModelingRA}'s multi-instance learning paradigm has inspired the development of methods that rank or select sentences from a bag of sentences that best represent a given relationship. For example, \citet{Zeng2015DistantSF} propose a method to score sentences within a bag based on their likelihood of expressing a relationship. \citet{lin-etal-2016-neural} introduced an attention mechanism that attends to relevant relational information within a bag of sentences. Advanced attention mechanisms were featured in numerous subsequent works \cite{Luo2017LearningWN, Han2018NeuralKA, Alt2019FinetuningPT}. 

However, not all works reported increased performance from applying an attention mechanism to a bag of sentences. On a distantly supervised dataset constructed by pairing PubMed abstracts \cite{canese2013pubmed} with the UMLS knowledge graph \cite{umls}, \citet{Amin2020ADA} reported a drop in RE performance when using selective attention compared to a simple average pooling over a bag of sentences. 

This phenomenon of poor results while using attention mechanisms with multi-instance learning on distantly supervised datasets was recently explored by \citet{Hu2021HowKG}. They show experimentally that the performance of attention mechanisms drops significantly compared to average pooling as the ratio of noise in a dataset increases. 

To conduct this experiment, the authors start with FewRel, a manually labeled sentence-level RE dataset, and inject noise into the dataset by randomly replacing entity pairs in a sentence with different entity pairs while keeping the context unchanged. They use this method to automatically construct nine datasets with varying ratios of noise. 

They report that, using multi-instance learning, simple average pooling outperforms attention mechanisms in all settings with noise. Without noise, attention mechanisms perform better than average pooling. They concluded that attention mechanisms applied to bags of sentences are not robust to noise. They posit that attention mechanisms learn to focus on the relevant context within a bag of sentences, but informative context is either rare or missing entirely within noisy bags of sentences. In such cases, better performance is obtained by only considering the entity mentions via average pooling and ignoring noisy context. 

Leveraging the knowledge that attention mechanisms and average pooling do well in different settings within multi-instance learning, \citet{Li2020SelfAttentionES} introduced a method that enables a model to learn when either method is most effective. The authors proposed a trainable selection gate that learns when to use attention versus average pooling for bags of sentences. The gate selects an attention mechanism to predict relationships for bags of sentences with good context (i.e., accurately labeled sentences). Likewise, the gate selects average pooling for bags of sentences with poor context (i.e., wrongly labeled, noisy sentences). The gating function is a simple sigmoid activation function that associates a low gating value for noisy sentences, effectively preventing noise propagation.

One shortcoming of multi-instance learning, in general, is its requirement of multiple sentences to fill a bag of sentences. In some domains, such as the biomedical domain, it can be rare to have particular entity pairs mentioned together in a sentence. Infrequently co-mentioned entity pairs lead to a long-tail distribution where a majority of entity pairs are supported by only one or two sentences \cite{Hogan2021AbstractifiedML}. Creating bags with these entity pairs requires heavy up-sampling, which diminishes the effectiveness of multi-instance learning. 

For example, using a bag size equal to 16 sentences and an entity pair that is supported by a single sentence, the single sentence is duplicated 15 times to fill the multi-instance learning bag. In the NYT10 dataset \cite{Riedel2010ModelingRA}, as many as 80\% of the bags of sentences contain a single duplicated sentence. A commonly mentioned entity pair may produce a bag of unique and informative sentences, while a rarely mentioned entity pair may produce a bag consisting of a single duplicated sentence. Furthermore, depending on the noise ratio in the dataset, a bag of sentences may not contain the context required to determine a relationship. 



\subsection{Pre-training Methods}
Current state-of-the-art relation extraction (RE) models typically leverage a two-phase training: a self-supervised pre-training followed by a supervised fine-tuning. This training methodology was popularized by BERT \cite{Devlin2019BERTPO} and has since made significant gains on the task of RE. To effectively extract relations from text, one may start with a pre-trained language model that comes with some latent semantic understanding (e.g., BERT, SciBERT \cite{Beltagy2019SciBERTAP}, PubMedBERT \cite{pubmed_bert}) and then use those informed representations to fine-tune a model to classify relationships between pairs of entities in text. 

However, one critique of pre-trained language models such as BERT and SciBERT is that they feature a generic pre-training objective. This is by design---the generic pre-training objective, namely masked language modeling, allows the model to generalize to a wide variety of downstream tasks such as text summarization, named entity recognition, question-answering, relation extraction, etc. 

\begin{figure}
    \centering
    \includegraphics[width=0.48\textwidth]{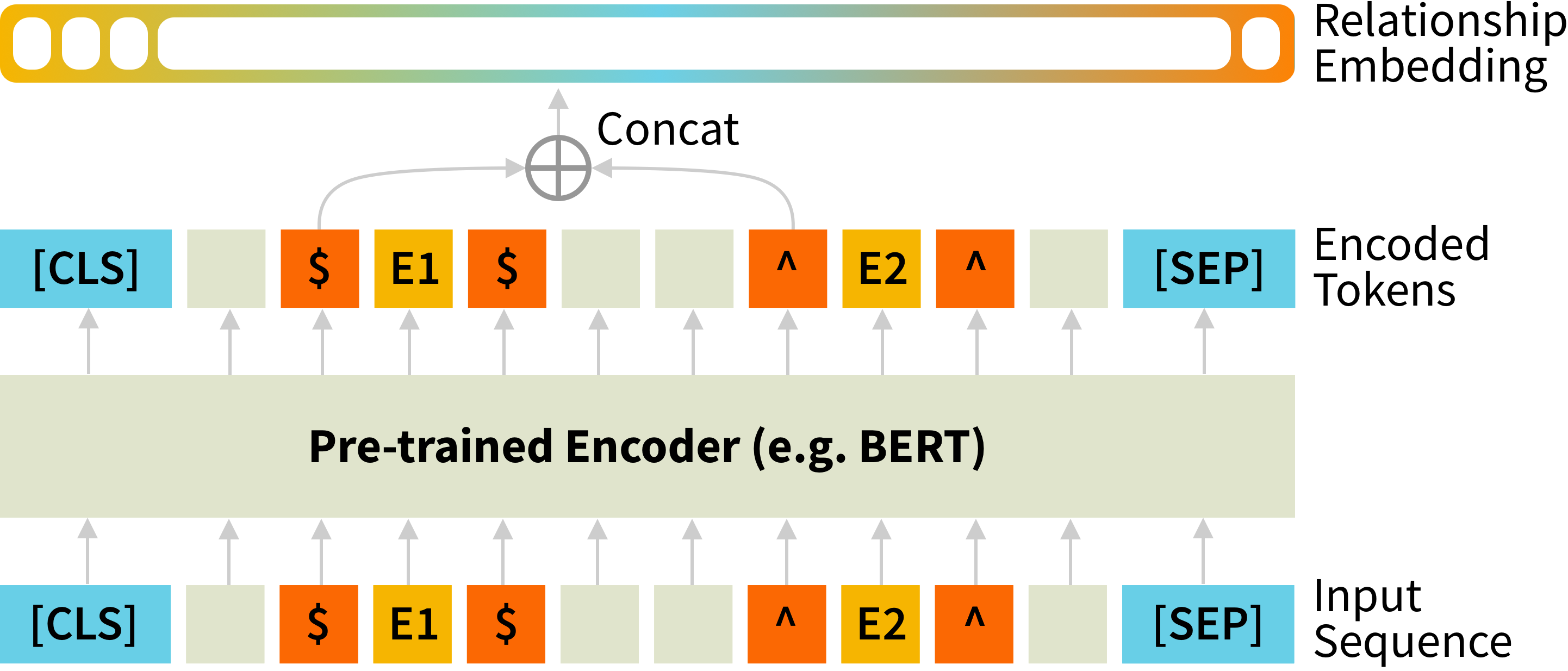}
    \caption{An illustration showing how \citet{BaldiniSoares2019MatchingTB} form relationship representations. Sentences with entities wrapped by special delimiter tokens (e.g., `\$' and `\^{}') are passed through BERT. Then, embeddings of the start tokens are concatenated to form the relationship embedding. During pre-training, a dot product calculates pairwise similarity between relationship embeddings.}
    \label{fig:re_embedding}
\end{figure}

Soon after BERT's release, \citet{BaldiniSoares2019MatchingTB} proposed ``Matching the Blanks'' (MTB), a model that featured a pre-training objective explicitly designed for the task of relation extraction. First, mentions within sentences are wrapped by special delimiter tokens marking their start and end spans, then  70\% of the entity mentions within a sentence are randomly masked. The modified input sequences with masked and delimited entities are sent through a BERT encoder to get an embedded sequence. Next, relationship representations are formed by concatenating the embedded start tokens of an entity pair (see Figure \ref{fig:re_embedding}). Finally, pairs relationship representations are assigned a similarity score via a dot product.

The model learns to ``match the blanks'' by assigning a high similarity between sentences that contain the same entity pair or, in other words, sentences that may express the same relationship. They empirically show that the delimiter tokens are essential to model performance. Without them, the model cannot decipher the entities of interest within a sentence. 

MTB learns the parameters for their relation statement encoder $f_{\theta}$---a binary classifier function that determines the similarity of two entity pairs:
$$
p\left(l=1 \mid \mathbf{r}, \mathbf{r}^{\prime}\right)=\frac{1}{1+\exp f_{\theta}(\mathbf{r})^{\top} f_{\theta}\left(\mathbf{r}^{\prime}\right)}
$$

Where $\mathbf{r}$ is the relationship representation formed using the process illustrated in Figure \ref{fig:re_embedding}. The inner product $\exp f_{\theta}(\mathbf{r})^{\top} f_{\theta}\left(\mathbf{r}^{\prime}\right)$ should be high if $\mathbf{r}$ and $\mathbf{r}^{\prime}$ express semantically similar relations and vice-versa. This binary classifier is used in the construction of their pre-training loss objective: 

$$
\begin{aligned}
\mathcal{L}(\mathcal{D})=-& \frac{1}{|\mathcal{D}|^{2}} \sum_{\left(\mathbf{r}, e_{1}, e_{2}\right) \in \mathcal{D}} \sum_{\left(\mathbf{r}^{\prime}, e_{1}^{\prime}, e_{2}^{\prime}\right) \in \mathcal{D}}
\\
& \delta_{e_{1}, e_{1}^{\prime}} \delta_{e_{2}, e_{2}^{\prime}} \cdot \log p\left(l=1 \mid \mathbf{r}, \mathbf{r}^{\prime}\right)+\\
&\left(1-\delta_{e_{1}, e_{1}^{\prime}} \delta_{e_{2}, e_{2}^{\prime}}\right) \\
&\cdot \log \left(1-p\left(l=1 \mid \mathbf{r}, \mathbf{r}^{\prime}\right)\right)
\end{aligned}
$$

Where $\delta_{e_{1}, e_{1}^{\prime}} = 1$  \textit{iff} ${e_{1} =e_{1}^{\prime}}$ and corpus of relation statements $\mathcal{D}=\left[\left(\mathbf{r}^{0}, e_{1}^{0}, e_{2}^{0}\right) \ldots\left(\mathbf{r}^{N}, e_{1}^{N}, e_{2}^{N}\right)\right]$. Each relational statement contains two entities $e_{1}^{i}$, $e_{2}^{i} \in \mathcal{E}$, where $\mathcal{E}$ represents the set of all entities. The corpus $\mathcal{D}$ used for pre-training consisted of 600 million relation statement pairs and was constructed by using Google Cloud Natural Language API \footnote{\url{https://cloud.google.com/natural-language}} to identify entities within passages from Wikipedia (English). 

Interestingly, the MTB model departs from the distantly supervised RE paradigm in that it does not use relation labels during the pre-training phase. They argue that this creates a more ontology-independent pre-trained model since training does not rely on an ontology-specific knowledge graph to define relational classes. 

In the second phase of MTB, the model is fine-tuned on an ontology-specific downstream task such as DocRED or FewRel. The authors report superior performance compared to generic pre-training models such as BERT. MTB achieves significant gains on low-resource training (e.g., using less than 20\% of the fine-tuning training data). Notably, MTB outperformed the best-published methods on FewRel without fine-tuning on FewRel data. This feat highlights MTB's effectiveness in learning high-quality representations for relationships.

MTB's success inspired numerous subsequent pre-trained models optimized for relation extraction. Continuing the thread of pre-trained models optimized for RE, \citet{xiao-etal-2020-denoising} propose a three-part pre-training objective optimized for document-level RE. They pre-train a model to match masked entities, identify relations, and align relational facts. 

For entity matching, they randomly mask entities within a document and have the model match the masked entities. This is similar to MTB but, instead of matching pairs of entities to other pairs of entities, they match single entities to each other. The authors claim that this helps the model learn anaphora resolution---resolving words referring to or replacing a word used earlier in a document. 

For relation detection, they use a binary classifier; relations are labeled as either a positive or a negative instance (e.g., ``no relation''). They show via an ablation experiment that relation detection is critical to the model’s overall performance as it helps the model reduce the noise introduced by distant supervision. Lastly, they use a pairwise similarity score to pairs of entities that express the same relationship to align relational facts.

They show that their three-part pre-trained model does well when evaluated on DocRED. It outperforms BERT as well as a few varieties of BERT (HIN-BERT \cite{Tang2020HINHI}, BERT-TS \cite{Wang2019FinetuneBF}), as well as a CNN, LSTM, and a BiLSTM.

Recent general-domain RE works have shown impressive performance gains using a contrastive learning pre-training objective designed specifically for relation extraction \cite{Peng2020LearningFC, Qin2021ERICAIE}. \citet{Peng2020LearningFC} propose a model named ``Contrastive Pre-training'' (CP), where they randomly mask entity mentions within a sentence and train the model to align different entity pairs that express the same relationship via contrastive learning. By masking entities, they encourage the model to learn from the context in a sentence instead of relying on entity mentions. They pre-train their model on a distantly supervised RE dataset constructed by pairing text from Wikipedia with the WikiData knowledge graph, mirroring the method used to construct the distantly labeled training set in DocRED. They then fine-tune CP on the TACRED dataset. The authors report that their model makes performance gains over BERT and MTB, showing that their contrastive learning framework effectively produces high-quality representations for relationships.

\begin{figure*}[t]
    \centering
    \includegraphics[width=1.0\textwidth]{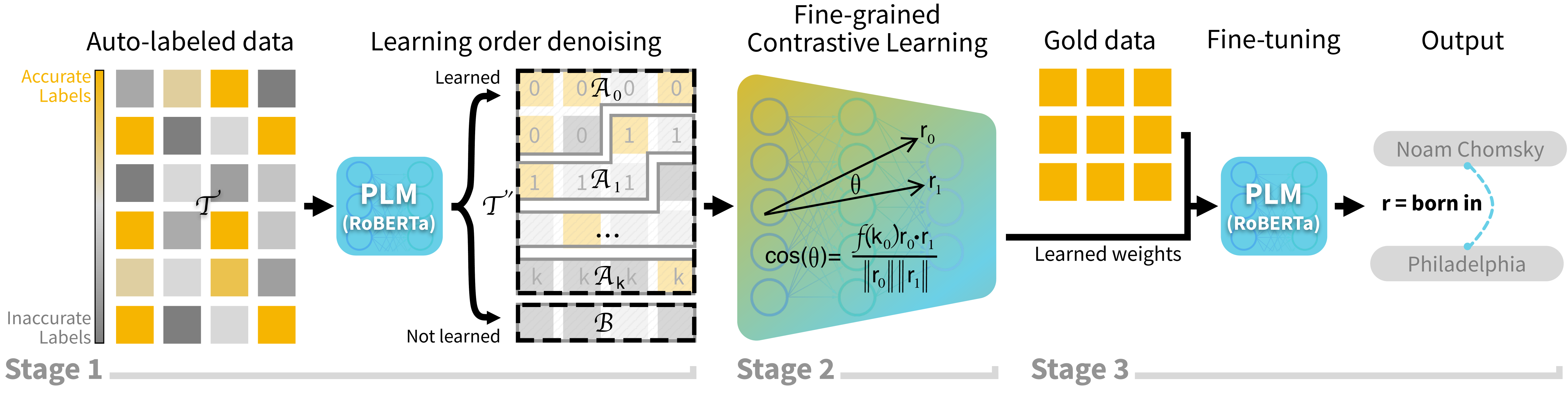}
    \caption{
    The FineCL \cite{hogan_fgcl} framework has three stages: 
    Stage 1: distantly supervised data ($\mathcal{T}$) is used to train a PLM via cross-entropy to collect ordered subsets of learned ($\mathcal{A}$) and not learned ($\mathcal{B}$) instances over $k$ epochs. Stage 2: function $f(k)$ weighs relation instances ($r_0, r_1$) relative to their learning order in a contrastive learning pre-training objective that uses cosine similarity to align similar relations. Stage 3: the model is adapted to a discriminative task.}
    \label{fig:fgcl}
\end{figure*}

The success of \citet{Peng2020LearningFC} inspired \citet{Qin2021ERICAIE}'s  ERICA model. ERICA uses a three-part loss objective for pre-training: entity discrimination, relation discrimination, and mask language modeling. For entity discrimination, they use contrastive learning to maximize the cosine similarity of a given head entity to a corresponding tail entity in a fact triple. Notably, they use document-level entity representations instead of sentence-level entity representations. Document-level entity representations average all mentions of a single entity within a document to form a representation. The authors claim this helps the model learn sentence-level as well as document-level relationships. 

For relation discrimination, they again use contrastive learning to group different instances of the same relationship. Finally, they include the masked language modeling objective introduced by BERT. To construct the pre-training dataset, they follow \citet{Peng2020LearningFC} and pair Wikipedia text with entities and relationships from the WikiData knowledge graph. The pre-trained ERICA model is then fine-tuned on numerous downstream tasks such as DocRED, TACRED, SemEval-2010 Task 8. They report improved performance in all settings compared to MTB and the method proposed by \citet{Peng2020LearningFC}, offering further evidence that contrastive learning for RE creates deep and informative relationship representations.

Conventional contrastive learning for RE treats all instances equally, which can be problematic when training on distantly supervised data that may contain up to 53\% of incorrectly assigned labels \cite{Gao2021ManualEM}. To diminish the impact of noise from automatically labeled relation data, \citet{hogan_fgcl} propose Fine-grained Contrastive Learning (FineCL). Figure \ref{fig:fgcl} illustrates the end-to-end data flow for the FineCL method. First, automatically labeled relation data is used to train an off-the-shelf language model (e.g., RoBERTa, or another domain-specific BERT relative) via cross-entropy. During training, the learning order of relation instances is recorded. A training instance is considered ``learned'' when the model first correctly predicts the corresponding instance. They show that the order of learned instances corresponds to label accuracy: clean, accurately labeled relation instances are, on average, learned first, followed by noisy, inaccurately labeled relation instances.

They then leverage the learning-order of relation instances to improve the relationship representations learned during pre-training by linearly projecting the weights of each relation instance corresponding to the order in which the instance was learned. Higher weights are applied to relation instances learned earlier in training relative to those learned later in training. These weights are used to inform a contrastive learning loss function that learns to group instances of similar relationships. They call this ``fine-grained'' contrastive learning since it leverages additional, fine-grained information about which instances are and are not noisy to produce higher-quality relationship representations. The representations learned during pre-training are then used to fine-tune the model on gold-labeled data.

%
%
\begin{table}[t]
\centering
\resizebox{\columnwidth}{!}{
\begin{tabular}{l | c c | c c | c c } 
    \hline
    \rule{0pt}{3ex}
    \textbf{Size}               & \multicolumn{2}{c}{1\%}   & \multicolumn{2}{c}{10\%}  & \multicolumn{2}{c}{100\%} \\  \hline
    \rule{0pt}{3ex}
    \textbf{Metrics}            & F1     & IgF1      & F1    & IgF1      & F1    & IgF1  \\ \hline 
    \rule{-2pt}{3ex}
    CNN                         & -      & -         & -     & -         & 42.3  & 40.3  \\ 
    BiLSTM                      & -      & -         & -     & -         & 51.1  & 50.3  \\  \hline   
    \rule{-3pt}{3ex}
    HINBERT                     & - & - & - & - & 55.6 & 53.7 \\
    CorefBERT                   & \underline{32.8} & \textbf{31.2} & 46.0 & 43.7 & 57.0 & 54.5 \\
    SpanBERT                    & 32.2 & 30.4 & 46.4 & 44.5 & 57.3 & 55.0 \\
    ERNIE                       & 26.7 & 25.5 & 46.7 & 44.2 & 56.6 & 54.2 \\
    MTB                         & 29.0 & 27.6 & 46.1 & 44.1 & 56.9 & 54.3 \\
    CP                          & 30.3 & 28.7 & 44.8 & 42.6 & 55.2 & 52.7 \\
    BERT                        & 19.9 & 18.8 & 45.2 & 43.1 & 56.6 & 54.4 \\
    RoBERTa                     & 29.6 & 27.9 & 47.6 & 45.7 & 58.2 & 55.9 \\
    ERICA\textsubscript{BERT}   & 22.9 & 21.7 & 48.5 & 46.4 & 57.4 & 55.2 \\
    ERICA\textsubscript{RoBERTa}& 30.0 & 28.2 & \underline{50.1} & \underline{48.1} & \underline{59.1} & \underline{56.9} \\ 
    FineCL                        & \textbf{33.2} & \textbf{31.2} & \textbf{50.3} & \textbf{48.3} & \textbf{59.5} & \textbf{57.1} \\ \hline
\end{tabular}} 
\caption{F1-micro scores on the DocRED test set as reported by \citet{hogan_fgcl}. IgF1 ignores performance on fact triples in the test set overlapping with triples in the train/dev sets.}
\label{tab:docred}
\end{table}

%
%
\begin{table}[t]
\centering
\resizebox{\columnwidth}{!}{
\begin{tabular}{l | c c c | c c c } 
    \hline
    \rule{0pt}{3ex}
    \textbf{Dataset}               & \multicolumn{3}{c}{TACRED}   & \multicolumn{3}{c}{SemEval} \\  \hline
    \rule{-3pt}{3ex}
    \textbf{Size}               & 1\%     & 10\%      & 100\%    & 1\%      & 10\%    & 100\%  \\ \hline 
    \rule{-3pt}{3ex}
    MTB                            & 35.7 & 58.8 & 68.2 & 44.2 & 79.2 & 88.2 \\
    CP                             & 37.1 & 60.6 & 68.1 & 40.3 & 80.0 & \underline{88.5} \\ 
    BERT                            & 22.2 & 53.5 & 63.7 & 41.0 & 76.5 & 87.8 \\
    RoBERTa                         & 27.3 & 61.1 & 69.3 & 43.6 & 77.7 & 87.5 \\
    ERICA\textsubscript{BERT}       & 34.9 & 56.0 & 64.9 & 46.4 & 79.8 & 88.1 \\
    ERICA\textsubscript{RoBERTa}    & \underline{41.1} & \underline{61.7} & \underline{69.5} & \underline{50.3} & \underline{80.9} & 88.4 \\ 
    FineCL                & \textbf{43.7} & \textbf{62.7} & \textbf{70.3} & \textbf{51.2} & \textbf{81.0} & \textbf{88.7}\\ \hline
\end{tabular}} 
\caption{F1-micro scores on the TACRED and SemEval test sets as reported by \citet{hogan_fgcl}.}
\label{tab:tacred_semeval}
\end{table}

Tables \ref{tab:docred} and  \ref{tab:tacred_semeval} compare FineCL's performance on DocRED, TACRED, and SemEval using the following aforementioned baselines: (1) CNN \cite{zeng-etal-2014-relation}, (2) BiLSTM \cite{lstm}, (3) BERT \cite{Devlin2019BERTPO}, (4) RoBERTa \cite{Liu2019RoBERTaAR}, (5) MTB \cite{BaldiniSoares2019MatchingTB}, (6) CP \cite{Peng2020LearningFC}, (7 \& 8) ERICA\textsubscript{BERT} \& ERICA\textsubscript{RoBERTa} \cite{Qin2021ERICAIE}. For fair comparison, each baseline is identically pre-trained mirroring the pre-training steps used in \citet{Qin2021ERICAIE}. In these experiments, \citet{hogan_fgcl} report performance gains over \citet{Qin2021ERICAIE} in all settings on DocRED, TACRED, and SemEval-2010 Task 8 datasets, illustrating the effectiveness of leveraging learning order to improve relation representations.
\section{Conclusion}
In this survey, we reviewed the relation extraction task with a focus on distant supervision. We traced the history of RE methods by discussing exemplary works and highlighting shortcomings to contextualize progress. We noted the key differences between pattern-based methods, statistical-based methods, neural-based methods, and finally, large language model-based methods. We conducted an overview of popular RE datasets and discussed some limitations of both corpus-based and instance-based RE evaluation. 

We reviewed current RE methods, focusing on denoising and pre-training methods---two of the defining methods of distant supervision for RE. We highlighted various pre-training methodologies used to attain state-of-the-art performance on RE benchmarks. Finally, we discussed recent RE works that leverage a contrastive learning pre-training objective.

\bibliography{bibliography}
\bibliographystyle{acl_natbib}
\end{document}